%% file: main.tex
\documentclass[11pt,letterpaper]{article}
\usepackage[nohyperref]{emnlp2017}
\usepackage{times}
\usepackage{latexsym}

\usepackage{url}
\usepackage{amsmath,amssymb}
\usepackage{graphicx}
\usepackage{color}
\graphicspath{ {figs/} }

\emnlpfinalcopy

\def\emnlppaperid{1296}

\title{Reinforced Video Captioning with Entailment Rewards}
\author{Ramakanth Pasunuru \and Mohit Bansal \\
  UNC Chapel Hill \\
  {\tt \{ram, mbansal\}@cs.unc.edu} \\
 }

\date{}

\begin{document}

\maketitle

\begin{abstract}

Sequence-to-sequence models have shown promising  improvements on the temporal task of video captioning, but they optimize word-level cross-entropy loss during training. 
First, using policy gradient and mixed-loss methods for reinforcement learning, 
we directly optimize sentence-level task-based metrics (as rewards), achieving significant improvements over the baseline, based on both automatic metrics and human evaluation on multiple datasets. Next, we propose a novel entailment-enhanced reward (CIDEnt) that corrects phrase-matching based metrics (such as CIDEr) to only allow for logically-implied partial matches and avoid contradictions, achieving further significant improvements over the CIDEr-reward model. Overall, our CIDEnt-reward model achieves the new state-of-the-art on the MSR-VTT dataset.
\end{abstract}

\input{introduction.tex}

\input{related_work.tex}

\input{model.tex}


\input{reward_functions.tex}

\input{results.tex}

\vspace{-5pt}
\section{Conclusion}
\vspace{-5pt}
\label{sec-conclusion}
We first presented a mixed-loss policy gradient approach for video captioning, allowing for metric-based optimization. We next presented an entailment-corrected CIDEnt reward that further improves results, achieving the new state-of-the-art on MSR-VTT. In future work, we are applying our entailment-corrected rewards
to other directed generation tasks such as image captioning and document summarization (using the new multi-domain NLI corpus~\cite{williams2017broad}).

\vspace{-4pt}
\section*{Acknowledgments}
\vspace{-5pt}
We thank the anonymous reviewers for their helpful comments. This work was supported by a Google Faculty Research Award, an IBM Faculty Award, a Bloomberg Data Science Research Grant, and NVidia GPU awards.

\appendix
\vspace{15pt}
\noindent
{\fontsize{12}{12}\selectfont \textbf{Supplementary Material}} \par
\input{appendix}

\bibliography{emnlp2017}
\bibliographystyle{emnlp_natbib}

\end{document}

%% file: introduction.tex
\section{Introduction}

\begin{figure}
\centering
\includegraphics[width=0.98\linewidth]{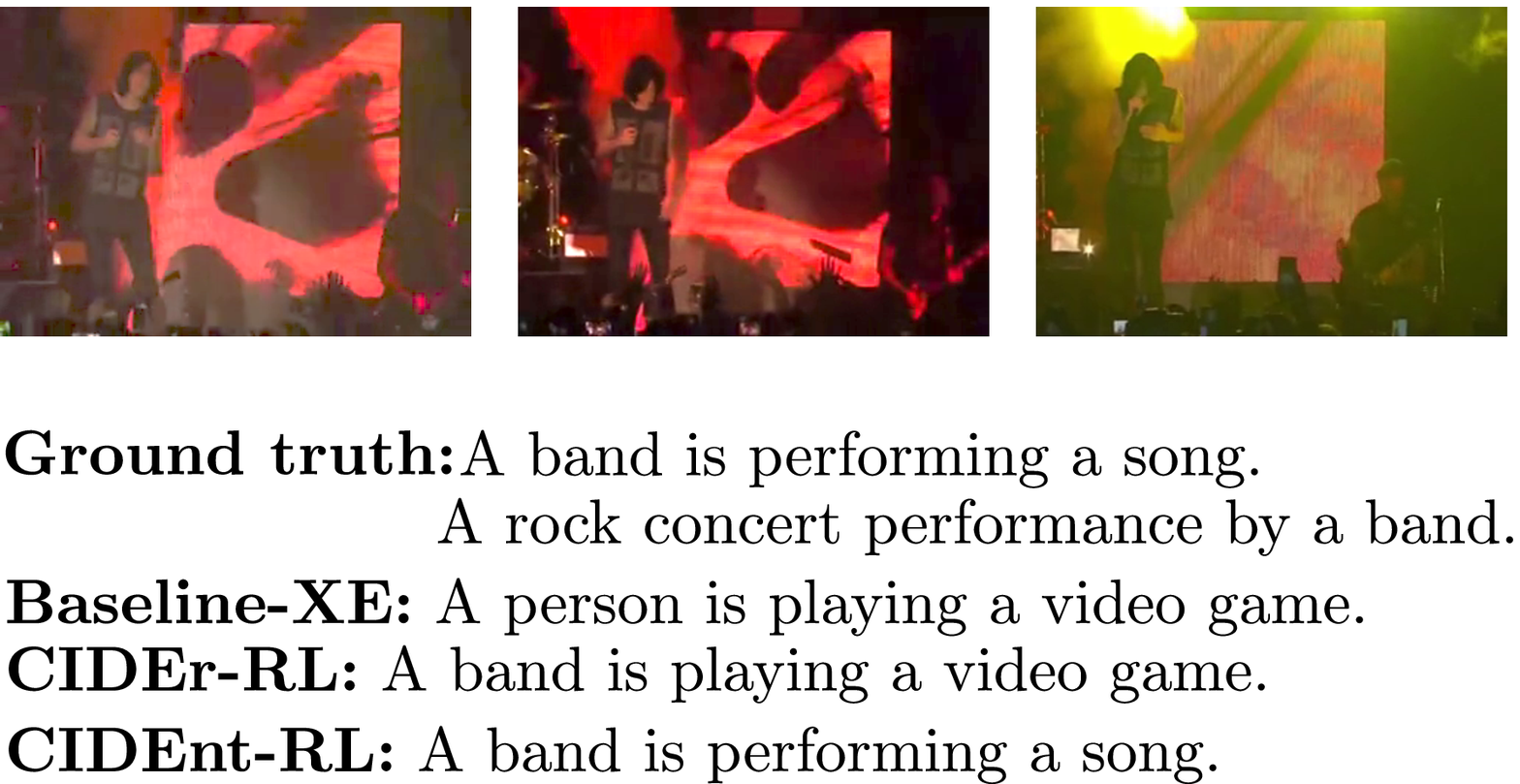}
\vspace{-10pt}
\caption{A correctly-predicted video caption generated by our CIDEnt-reward model.
}
\vspace{-10pt}
\label{fig:introexample}
\end{figure}

The task of video captioning (Fig.~\ref{fig:introexample}) is an important next step to image captioning, with additional modeling of temporal knowledge and action sequences, and has several applications in online content search, assisting the visually-impaired, etc.
Advancements in neural sequence-to-sequence learning has shown promising improvements on this task, based on encoder-decoder, attention, and hierarchical models~\cite{venugopalan2015sequence,pan2015hierarchical}.
However, these models are still trained using a word-level cross-entropy loss, which does not correlate well with the sentence-level metrics that the task is finally evaluated on (e.g., CIDEr, BLEU).
Moreover, these models suffer from exposure bias~\cite{ranzato2015sequence}, which occurs when a model is only exposed to the training data distribution, instead of its own predictions.
First, using a sequence-level training, policy gradient approach~\cite{ranzato2015sequence}, we allow video captioning models to directly optimize these non-differentiable metrics, as rewards in a reinforcement learning paradigm. We also address the exposure bias issue by using a mixed-loss~\cite{paulus2017deep,wu2016google}, i.e., combining the
cross-entropy and reward-based losses, which also helps maintain output fluency.

\begin{figure*}
\centering
\includegraphics[width=0.85\linewidth]{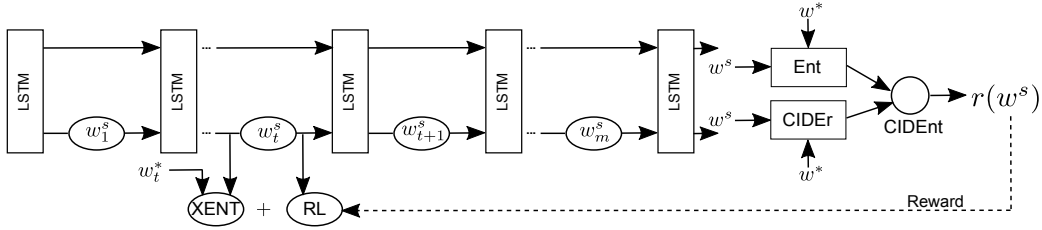}
\vspace{-10pt}
\caption{Reinforced (mixed-loss) video captioning using entailment-corrected CIDEr score as reward.}
\vspace{-8pt}
\label{fig:model}
\end{figure*}

Next, we introduce a novel entailment-corrected reward that checks for logically-directed partial matches. Current reinforcement-based text generation works use traditional phrase-matching metrics (e.g., CIDEr, BLEU) as their reward function. However, these metrics use \emph{undirected} $n$-gram matching of the machine-generated caption with the ground-truth caption, and hence fail to capture its \emph{directed logical correctness}. Therefore, they still give high scores to even those generated captions that contain a single but critical wrong word (e.g., negation, unrelated action or object), because all the other words still match with the ground truth.
We introduce CIDEnt, which penalizes the phrase-matching metric (CIDEr) based reward, when the entailment score is low. 
This ensures that a generated caption gets a high reward only when it is a directed match with (i.e., it is logically implied by) the ground truth caption, hence avoiding contradictory or unrelated information (e.g., see Fig.~\ref{fig:introexample}). 
Empirically, we show that first the CIDEr-reward model achieves significant improvements over the cross-entropy baseline (on multiple datasets, and automatic and human evaluation); next, the CIDEnt-reward model further achieves significant improvements over the CIDEr-based reward. Overall, we achieve the new state-of-the-art on the MSR-VTT dataset.

%% file: related_work.tex
\begin{table*}
\begin{center}
\begin{small}
\begin{tabular}{|l|l|c|c|}
\hline
Ground-truth caption & Generated (sampled) caption & CIDEr & Ent \\
\hline
a man is spreading some butter in a pan & puppies is melting butter on the pan &140.5&0.07 \\
a panda is eating some bamboo&a panda is eating some fried&256.8& 0.14\\
a monkey pulls a dogs tail&a monkey pulls a woman&116.4&0.04\\
a man is cutting the meat&a man is cutting meat into potato&114.3&0.08\\
the dog is jumping in the snow&a dog is jumping in cucumbers&126.2&0.03\\
a man and a woman is swimming in the pool&a man and a whale are swimming in a pool&192.5&0.02\\
\hline
\end{tabular}
\end{small}
\end{center}
\vspace{-10pt}
\caption{Examples of captions sampled during policy gradient and their CIDEr vs Entailment scores.}
\vspace{-8pt}
\label{table:cident}
\end{table*}

\vspace{-3pt}
\section{Related Work}
\vspace{-3pt}
Past work has presented several sequence-to-sequence models for video captioning, using attention, hierarchical RNNs, 3D-CNN video features, joint embedding spaces, language fusion, etc., but using word-level cross entropy loss training~\cite{venugopalan2015sequence,yao2015describing,pan2015hierarchical,pan2015jointly,venugopalan2016improving}.

Policy gradient for image captioning was recently presented by~\newcite{ranzato2015sequence}, using a mixed sequence level training paradigm to use non-differentiable evaluation metrics as rewards.\footnote{Several papers have presented the relative comparison of image captioning metrics, and their pros and cons~\cite{vedantam2015cider,anderson2016spice,liu2016improved,hodosh2013framing,elliott2014comparing}.}
\newcite{liu2016improved} and~\newcite{rennie2016self} improve upon this using Monte Carlo roll-outs and a test inference baseline, respectively. 
\newcite{paulus2017deep} presented summarization results with ROUGE rewards, in a mixed-loss setup.

Recognizing Textual Entailment (RTE) is a traditional NLP task~\cite{dagan2006pascal,lai2014illinois,jimenez2014unal}, boosted by a large dataset (SNLI) recently introduced by~\newcite{bowman2015large}. There have been several leaderboard models on SNLI~\cite{cheng2016long,rocktaschel2015reasoning}; we focus on the decomposable, intra-sentence attention model of~\newcite{parikh2016decomposable}.
Recently,~\newcite{pasunuru2017multitask} used multi-task learning to combine video captioning with entailment and video generation.

%% file: model.tex
\vspace{-3pt}
\section{Models}
\vspace{-3pt}

\paragraph{Attention Baseline (Cross-Entropy)}
Our attention-based seq-to-seq baseline model is similar to the~\newcite{bahdanau2014neural} architecture, where we encode input frame level video features $\{f_{1:n}\}$ via a bi-directional LSTM-RNN and then generate the caption $w_{1:m}$ using an LSTM-RNN with an attention mechanism. 
Let $\theta$ be the model parameters and $w^*_{1:m}$ be the ground-truth caption, then the cross entropy loss function is:
\begin{equation}
L(\theta) = -\sum_{t=1}^{m} \log p(w^*_t|w^*_{1:t-1},f_{1:n})
\end{equation}
where $p(w_t|w_{1:t-1},{f_{1:n}}) = softmax(W^T h^d_t)$,
 $W^T$ is the projection matrix, and $w_t$ and $h^d_{t}$ are the generated word and the RNN decoder hidden state at time step $t$, computed using the standard RNN recursion and attention-based context vector $c_t$. Details of the attention model are in the supplementary (due to space constraints). 

\paragraph{Reinforcement Learning (Policy Gradient)}
In order to directly optimize the sentence-level test metrics (as opposed to the cross-entropy loss above), we use a policy gradient $p_\theta$, where $\theta$ represent the model parameters. Here, our baseline model acts as an agent and interacts with its environment (video and caption). At each time step, the agent generates a word (action), and the generation of the end-of-sequence token results in a reward $r$ to the agent. Our training objective is to minimize the negative expected reward function:
\begin{equation}
L(\theta) =  - \mathbb{E}_{w^s \sim p_\theta}[r(w^s)]
\end{equation}

where $w^s$ is the word sequence sampled from the model.
Based on the REINFORCE algorithm~\cite{williams1992simple}, the gradients of this non-differentiable, reward-based loss function are:
\begin{equation}
\nabla_\theta L(\theta)  =  - \mathbb{E}_{w^s \sim p_\theta}[r(w^s) \cdot \nabla_\theta \log p_\theta(w^s)]
\end{equation}
We follow~\newcite{ranzato2015sequence}  approximating the above gradients via a single sampled word sequence. We also use a variance-reducing bias (baseline) estimator in the reward function. Their details and the partial derivatives using the chain rule are described in the supplementary.

\paragraph{Mixed Loss}
During reinforcement learning, optimizing for only the reinforcement loss (with automatic metrics as rewards) doesn't ensure the readability and fluency of the generated caption, and there is also a chance of gaming the metrics without actually improving the quality of the output~\cite{liu2016not}. Hence, for training our reinforcement based policy gradients, we use a mixed loss function, which is a weighted combination of the cross-entropy loss (XE) and the reinforcement learning loss (RL), similar to the previous work~\cite{paulus2017deep,wu2016google}. This mixed loss improves results on the metric used as reward through the reinforcement loss (and improves relevance based on our entailment-enhanced rewards) but also ensures better readability and fluency due to the cross-entropy loss (in which the training objective is a conditioned language model, learning to produce fluent captions). Our mixed loss is defined as:
\vspace{-5pt}
\begin{equation}
L_{\textsc{MIXED}} = (1-\gamma) L_{\textsc{XE}} + \gamma L_{\textsc{RL}}
\vspace{-5pt}
\end{equation}
where $\gamma$ is a tuning parameter used to balance the two losses. For annealing and faster convergence, we start with the optimized cross-entropy loss baseline model, and then move to optimizing the above mixed loss function.\footnote{We also experimented with the curriculum learning `MIXER' strategy of~\newcite{ranzato2015sequence}, where the XE+RL annealing is based on the decoder time-steps; however, the mixed loss function strategy (described above) performed better in terms of maintaining output caption fluency.}

%% file: reward_functions.tex
\section{Reward Functions}
\label{sec:entreward}

\noindent\textbf{Caption Metric Reward} \hspace{3pt} Previous image captioning papers have used traditional captioning metrics such as CIDEr, BLEU, or METEOR as reward functions, based on the match between the generated caption sample and the ground-truth reference(s). First, it has been shown by~\newcite{vedantam2015cider} that CIDEr, based on a consensus measure across several human reference captions, has a higher correlation with human evaluation than other metrics such as METEOR, ROUGE, and BLEU. They further showed that CIDEr gets better with more number of human references (and this is a good fit for our video captioning datasets, which have 20-40 human references per video). 

More recently,~\newcite{rennie2016self} further showed that CIDEr as a reward in image captioning outperforms all other metrics as a reward, not just in terms of improvements on CIDEr metric, but also on all other metrics. In line with these above previous works, we also found that CIDEr as a reward (`CIDEr-RL' model) achieves the best metric improvements in our video captioning task, and also has the best human evaluation improvements (see Sec.~\ref{sec:othermetrics} for result details, incl. those about other rewards based on BLEU, SPICE).

\vspace{5pt}
\noindent\textbf{Entailment Corrected Reward} \hspace{1pt} 
Although CIDEr performs better than other metrics as a reward, all these metrics (including CIDEr) are still based on an undirected $n$-gram matching score between the generated and ground truth captions.
For example, the wrong caption ``a man is playing football'' w.r.t. the correct caption ``a man is playing basketball'' still gets a high score, even though these two captions belong to two completely different events. Similar issues hold in case of a negation or a wrong action/object in the generated caption (see examples in Table~\ref{table:cident}).

We address the above issue by using an entailment score to correct the phrase-matching metric (CIDEr or others) when used as a reward, ensuring that the generated caption is logically implied by (i.e., is a paraphrase or directed partial match with) the ground-truth caption.
To achieve an accurate entailment score, we adapt the state-of-the-art decomposable-attention model of~\newcite{parikh2016decomposable} trained on the SNLI corpus (image caption domain). This model gives us a probability for whether the sampled video caption (generated by our model) is entailed by the ground truth caption as premise (as opposed to a contradiction or neutral case).\footnote{Our entailment classifier based on~\newcite{parikh2016decomposable} is 92\% accurate on entailment in the caption domain, hence serving as a highly accurate reward score. For other domains in future tasks such as new summarization, we plan to use the new multi-domain dataset by~\newcite{williams2017broad}.}   
Similar to the traditional metrics, the overall `Ent' score is the maximum over the entailment scores for a generated caption w.r.t. each reference human caption (around 20/40 per MSR-VTT/YouTube2Text video).
CIDEnt is defined as:
\vspace{-1pt}
\begin{equation}
\text{CIDEnt} = 
\begin{cases}
   		\text{CIDEr}-\lambda,   & \text{if } \;\; \text{Ent} < \beta \\
   		\text{CIDEr},              & \text{otherwise}
\end{cases} 
\end{equation}
which means that if the entailment score is very low, we  penalize the metric reward score by decreasing it by a penalty $\lambda$.  
This agreement-based formulation ensures that we only trust the CIDEr-based reward in cases when the entailment score is also high. Using CIDEr$-\lambda$ also ensures the smoothness of the reward w.r.t. the original CIDEr function (as opposed to clipping the reward to a constant). 
Here, $\lambda$ and $\beta$ are hyperparameters that can be tuned on the dev-set; on light tuning, we found the best values to be intuitive: $\lambda = $ roughly the baseline (cross-entropy) model's score on that metric (e.g., $0.45$ for CIDEr on MSR-VTT dataset); and $\beta = 0.33$ (i.e., the 3-class entailment classifier chose contradiction or neutral label for this pair). 
Table~\ref{table:cident} shows some examples of sampled generated captions during our model training, where CIDEr was misleadingly high for incorrect captions, but the low entailment score (probability) helps us successfully identify these cases and penalize the reward.

%% file: results.tex
\vspace{-5pt}
\section{Experimental Setup}
\vspace{-5pt}

\noindent\textbf{Datasets} 
We use 2 datasets: MSR-VTT~\cite{xu2016msr} has $10,000$ videos, $20$ references/video; and YouTube2Text/MSVD~\cite{chen2011collecting} has $1970$ videos, $40$ references/video. Standard splits and other details in supp.

\noindent\textbf{Automatic Evaluation}
We use several standard automated evaluation metrics: METEOR, BLEU-4, CIDEr-D, and ROUGE-L (from MS-COCO evaluation server~\cite{chen2015microsoft}).

\noindent\textbf{Human Evaluation}
\label{subsubsec-human_eval}
We also present human evaluation for comparison of baseline-XE, CIDEr-RL, and CIDEnt-RL models, esp. because the automatic metrics cannot be trusted solely. Relevance measures how related is the generated caption w.r.t,  to the video content, whereas coherence measures readability of the generated caption.

\noindent\textbf{Training Details}
All the hyperparameters are tuned on the validation set. All our results (including baseline) are based on a 5-avg-ensemble. See supplementary for extra training details, e.g., about the optimizer, learning rate, RNN size, Mixed-loss, and CIDEnt hyperparameters.

\begin{table*}
\begin{center}
\begin{small}
\begin{tabular}{|l|c|c|c|c|c||c|}
\hline
Models & BLEU-4 & METEOR & ROUGE-L & CIDEr-D & CIDEnt  & Human*\\
\hline
\multicolumn{7}{|c|}{\textsc{Previous Work}}\\
\hline
Venugopalan ~\shortcite{venugopalan2014translating}$^\star$ & 32.3 & 23.4 & - & - & - & -\\
\newcite{yao2015describing}$^\star$ & 35.2 & 25.2 & - & - & - & -\\
\newcite{xu2016msr} & 36.6 & 25.9 & - & - & - & -\\
\newcite{pasunuru2017multitask} & \textbf{40.8} & \textbf{28.8} &60.2 &47.1 &-&-\\
\hline
Rank1: v2t\_navigator & \textbf{40.8} & 28.2 & 60.9 & 44.8  & - & -\\
Rank2: Aalto & 39.8 & 26.9  & 59.8 & 45.7 & - & - \\
Rank3: VideoLAB & 39.1 & 27.7  & 60.6 & 44.1 & - & -\\
\hline
\multicolumn{7}{|c|}{\textsc{Our Models}}\\
\hline
Cross-Entropy (Baseline-XE) & 38.6 & 27.7 & 59.5 & 44.6 & 34.4 & - \\
CIDEr-RL & 39.1 & 28.2 & 60.9 & 51.0 & 37.4 & 11.6 \\
CIDEnt-RL (\textbf{New Rank1}) & \textbf{40.5} & \textbf{28.4} & \textbf{61.4}  & \textbf{51.7} &\textbf{44.0}& \textbf{18.4} \\
\hline
\end{tabular}
\end{small}
\end{center}
\vspace{-10pt}
\caption{Our primary video captioning results on MSR-VTT. All CIDEr-RL results are statistically significant over the baseline XE results, and all CIDEnt-RL results are stat. signif. over the CIDEr-RL results. Human* refers to the `pairwise' comparison of human relevance evaluation between CIDEr-RL and CIDEnt-RL models (see full human evaluations of the 3 models in Table~\ref{table-humanevaluationbaseline} and Table~\ref{table-humanevaluation}).
}
\vspace{-3pt}
\label{table-baseandmultitaskresults}
\end{table*}

\vspace{-5pt}
\section{Results}
\vspace{-5pt}
\subsection{Primary Results}
Table~\ref{table-baseandmultitaskresults} shows our primary results on the popular MSR-VTT dataset.
First, our baseline attention model trained on cross entropy loss (`Baseline-XE') achieves strong results w.r.t. the previous state-of-the-art methods.\footnote{We list previous works' results as reported by the MSR-VTT dataset paper itself, as well as their 3 leaderboard winners ({\scriptsize\url{http://ms-multimedia-challenge.com/leaderboard}}), plus the 10-ensemble video+entailment generation  multi-task model of~\newcite{pasunuru2017multitask}.} Next, our policy gradient based mixed-loss RL model with reward as CIDEr (`CIDEr-RL') improves significantly\footnote{Statistical significance of $p < 0.01$ for CIDEr, METEOR, and ROUGE, and $p < 0.05$ for BLEU, based on the bootstrap test~\cite{noreen1989computer,efron1994introduction}.} over the baseline on all metrics, and not just the CIDEr metric. It also achieves statistically significant improvements in terms of human relevance evaluation (see below).
Finally, the last row in Table~\ref{table-baseandmultitaskresults} shows results for our novel CIDEnt-reward RL model (`CIDEnt-RL'). This model achieves statistically significant\footnote{Statistical significance of $p < 0.01$ for CIDEr, BLEU, ROUGE, and CIDEnt, and $p < 0.05$ for METEOR.} improvements on top of the strong CIDEr-RL model, on all automatic metrics (as well as human evaluation).
Note that in Table~\ref{table-baseandmultitaskresults}, we also report the CIDEnt reward scores, and the CIDEnt-RL model strongly outperforms CIDEr and baseline models on this entailment-corrected measure. Overall, we are also the new Rank1 on the MSR-VTT leaderboard, based on their ranking criteria.

\paragraph{Human Evaluation}
We also perform small human evaluation studies (250 samples from the MSR-VTT test set output) to compare our 3 models pairwise.\footnote{We randomly shuffle pairs to anonymize model identity and the human evaluator then chooses the better caption based on relevance and coherence (see Sec.~\ref{subsubsec-human_eval}). `Not Distinguishable' are cases where the annotator found both captions to be equally good or equally bad).} As shown in Table~\ref{table-humanevaluationbaseline} and Table~\ref{table-humanevaluation}, in terms of relevance, first our CIDEr-RL model stat. significantly outperforms the baseline XE model ($p<0.02$); next, our CIDEnt-RL model significantly outperforms the CIDEr-RL model ($p<0.03$). The models are statistically equal on coherence in both comparisons.


\begin{table}[t]
\begin{small}
\begin{center}
\begin{tabular}{l  c  c }
\hline
& Relevance & Coherence \\
\hline
 Not  Distinguishable & 64.8\% & 92.8\%  \\
 Baseline-XE Wins & 13.6\% & 4.0\%  \\
 CIDEr-RL Wins & \textbf{21.6\%} & 3.2\%  \\
\hline
\end{tabular}
\end{center}
\end{small}
\vspace{-12pt}
\caption{Human eval: Baseline-XE vs CIDEr-RL.}
\vspace{2pt}
\label{table-humanevaluationbaseline}
\end{table}

\begin{table}[t]
\begin{small}
\begin{center}
\begin{tabular}{l  c  c }
\hline
& Relevance & Coherence \\
\hline
 Not  Distinguishable & 70.0\% & 94.6\%  \\
 CIDEr-RL Wins & 11.6\% & 2.8\%  \\
 CIDEnt-RL Wins & \textbf{18.4\%} & 2.8\%  \\
\hline
\end{tabular}
\end{center}
\end{small}
\vspace{-10pt}
\caption{Human eval: CIDEr-RL vs CIDEnt-RL.}
\vspace{-7pt}
\label{table-humanevaluation}
\end{table}

\vspace{-5pt}
\subsection{Other Datasets}
\vspace{-2pt}
We also tried our CIDEr and CIDEnt reward models on the YouTube2Text dataset. In Table~\ref{table-msvdresults}, we first see strong improvements from our CIDEr-RL model on top of the cross-entropy baseline. Next, the CIDEnt-RL model also shows some improvements over the CIDEr-RL model, e.g., on BLEU and the new entailment-corrected CIDEnt score. It also achieves significant improvements on human relevance evaluation (250 samples).\footnote{This dataset has a very small dev-set, causing tuning issues -- we plan to use a better train/dev re-split in future work.}


\begin{table}[t]
\begin{small}
\begin{center}
\begin{tabular}{|l|c|c|c|c|c||c|}
\hline
Models & B & M & R & C & CE & H*\\
\hline
{\scriptsize Baseline-XE} & 52.4 & 35.0 & 71.6 & 83.9 & 68.1 & -\\
{\scriptsize CIDEr-RL} & 53.3  & 35.1 & 72.2 &  89.4 & 69.4 & 8.4 \\
{\scriptsize CIDEnt-RL} & 54.4 & 34.9 & 72.2 & 88.6 & 71.6 & 13.6 \\
\hline
\end{tabular}
\vspace{-4pt}
\caption{Results on YouTube2Text (MSVD) dataset. CE = CIDEnt score. H* refer to the pairwise human comparison of relevance.}
\vspace{-10pt}
\label{table-msvdresults}
\end{center}
\end{small}
\end{table}

\vspace{-5pt}
\subsection{Other Metrics as Reward}
\vspace{-2pt}
\label{sec:othermetrics}
As discussed in Sec.~\ref{sec:entreward}, CIDEr is the most promising metric to use as a reward for captioning, based on both previous work's findings as well as ours. We did investigate the use of other metrics as the reward. When using BLEU as a reward (on MSR-VTT), we found that this BLEU-RL model achieves BLEU-metric improvements, but was worse than the cross-entropy baseline on human evaluation. Similarly, a BLEUEnt-RL model achieves BLEU and BLEUEnt metric improvements, but is again worse on human evaluation. We also experimented with the new SPICE metric~\cite{anderson2016spice} as a reward, but this produced long repetitive phrases (as also discussed in~\newcite{liu2016improved}).

\vspace{-5pt}
\subsection{Analysis}
\vspace{-3pt}
Fig.~\ref{fig:introexample} shows an example where our CIDEnt-reward model correctly generates a ground-truth style caption, whereas the CIDEr-reward model produces a non-entailed caption because this caption will still get a high phrase-matching score. Several more such examples are in the supp.

%% file: appendix.tex
\section{Attention-based Baseline Model (Cross-Entropy)}
Our attention baseline model is similar to the~\newcite{bahdanau2014neural} architecture, where  we encode input frame level video features to a bi-directional LSTM-RNN and then generate the caption using a single layer LSTM-RNN, with an attention mechanism. Let $\{f_1,f_2,...,f_n\}$ be the frame-level features of a video clip and $\{w_1,w_2,...,w_m\}$ be the sequence of words forming a caption. The distribution of words at time step $t$ given the previously generated words and input video frame-level features is given as follows:
\begin{equation}
p(w_t|w_{1:t-1},{f_{1:n}}) = softmax(W^T h^d_t)
\end{equation}
where $w_t$ and $h^d_{t}$ are the generated word and hidden state of the LSTM decoder at time step $t$. $W^T$ is the projection matrix. $h^d_t$ is given as follows:
\begin{equation}
h^d_t = S(h^d_{t-1},w_{t-1},c_t)
\end{equation}
where $S$ is a non-linear function. $h^d_{t-1}$ and $w_{t-1}$ are  previous time step's hidden state and generated word.  $c_t$ is a context vector which  is a linear weighted combination of the encoder hidden states $h^e_i$, given by $c_t = \sum \alpha_{t,i}h^e_i$. These weights $\alpha_{t,i}$ act as an attention mechanism, and are defined as follows:
\begin{equation}
\alpha_{t,i} =  \frac{exp(e_{t,i})}{\sum_{k=1}^n exp(e_{t,k})}
\end{equation}
where the attention function $e_{t,i}$ is defined as:
\begin{equation}
e_{t,i} =  w^T tanh(W_a h^e_i + U_a h^d_{t-1}+ b_a)
\end{equation}
where $w$, $W_a$, $Ua$, and $b_a$ are trained attention parameters. Let $\theta$ be the model parameters and $\{w^*_1,w^*_2,...,w^*_m\}$ be the ground-truth word sequence, then the cross entropy loss optimization function is defined as follows:
\begin{equation}
L(\theta) = -\sum_{i=1}^{m} \log (p(w^*_t|w^*_{1:t-1},f_{1:n}))
\end{equation}
\section{Reinforcement Learning (Policy Gradient)}
Traditional video captioning systems minimize the cross entropy loss during training, but typically evaluated using phrase-matching metrics: BLEU, METEOR, CIDEr, and ROUGE-L.
This discrepancy can be addressed by directly optimizing the non-differentiable metric scores using policy gradients $p_\theta$, where $\theta$ represents the model parameters. In our captioning system, our baseline attention model acts as an agent and interacts with its environment (video and caption). At each time step, the agent generates a word (action), and the generation of the end-of-sequence token results in a reward $r$ to the agent. Our training objective is to minimize the negative expected reward function given by:
\begin{equation}
L(\theta) =  - \mathbb{E}_{w^s \sim p_\theta}[r(w^s)]
\end{equation}
where $w^s=\{w^s_1,w^s_2,...,w^s_m\}$, and $w^s_t$ is the word sampled from the model at time step $t$.
Based on the REINFORCE algorithm~\cite{williams1992simple}, the gradients of the non-differentiable, reward-based loss function can be computed as follows:
\begin{equation}
\nabla_\theta L(\theta)  =  - \mathbb{E}_{w^s \sim p_\theta}[r(w^s) \nabla_\theta \log p_\theta(w^s)]
\end{equation}

The above gradients can be approximated from a single sampled word sequence $w^s$ from $p_\theta$ as follows:
\begin{equation}
\nabla_\theta L(\theta)  \approx - r(w^s) \nabla_\theta \log p_\theta(w^s)
\label{eq-sampled_gradient}
\end{equation}

However, the above approximation has high variance because of estimating the gradient with a single sample. Adding a baseline estimator reduces this variance~\cite{williams1992simple} without changing the expected gradient.
Hence, Eqn:~\ref{eq-sampled_gradient} can be rewritten as follows:
\begin{equation}
\nabla_\theta L(\theta)  \approx - (r(w^s)-b_t) \nabla_\theta \log p_\theta(w^s)
\end{equation}
where $b_t$ is the baseline estimator, where $b_t$ can be a function of $\theta$ or time step $t$, but not a function of $w^s$. In our model, baseline estimator is a simple linear regressor with hidden state of the decoder $h^d_t$ at time step $t$ as the input. We stop the back propagation of gradients before the hidden states for the baseline bias estimator.  Using the chain rule, loss function can be written as:
\begin{equation}
\nabla_\theta L(\theta) = \sum_{t=1}^m \frac{\partial L}{\partial s_t} \frac{\partial s_t}{\partial \theta}
\end{equation}
where $s_t$ is the input to the \emph{softmax} layer, where $s_t=W^T h^d_t$.  $\frac{\partial L}{\partial s_t}$ is given by~\cite{zaremba2015reinforcement} as follows:
\begin{equation}
\frac{\partial L}{\partial s_t} \approx (r(w^s)-b_t) (p_\theta(w_t|h^d_t) - 1_{w^s_t})
\end{equation}

The overall intuition behind this gradient formulation is: if the reward $r(w^s)$ for the sampled word sequence $w^s$ is greater than the baseline estimator $b_t$, the gradient of the loss function becomes negative, then model encourages the sampled distribution by increasing their word probabilities, otherwise the model discourages the sampled distribution by decreasing their word probabilities.

\section{Experimental Setup}
\subsection{MSR-VTT Dataset}
MSR-VTT  is a  diverse  collection of $10,000$ video clips (41.2 hours of duration) from  a commercial video search engine.  Each video has $20$ human annotated reference captions collected through Amazon Mechanical Turk (AMT).  We use the standard split as provided in~\cite{xu2016msr}, i.e., $6513$ for training, $497$ for testing , and remaining for testing. For each video, we sample at $3 fps$ and we extract Inception-v4~\cite{szegedy2016inception} features from these sampled frames and we also remove all the punctuations from the text data.

\subsection{YouTube2Text Dataset} 
We also evaluate our models on YouTube2Text dataset~\cite{chen2011collecting}. This dataset has $1970$ video clips and each clip is annotated with an average of $40$ captions by humans. We use the standard split as given in~\cite{venugopalan2015sequence}, i.e., $1200$ clips for training, $100$ for validation and $670$ for testing. We do similar pre-processing as the MSR-VTT dataset.

\subsection{Automatic Evaluation Metrics}
We use several standard automated evaluation metrics: METEOR~\cite{banerjee2005meteor}, BLEU-4~\cite{papineni2002bleu}, CIDEr-D~\cite{vedantam2015cider}, and ROUGE-L~\cite{lin2004rouge}. We use the standard Microsoft-COCO evaluation server~\cite{chen2015microsoft}.

\subsection{Human Evaluation}
Apart from the automatic metrics, we also present human evaluation comparing the CIDEnt-reward model with the CIDEr-reward model, esp. because the automatic metrics cannot be trusted solely. Human evaluation uses \emph{Relevance} and \emph{Coherence} as the comparison metrics. Relevance is about  how related is the generated caption w.r.t. the content of the video, whereas coherence is about the logic, fluency, and readability of the generated caption.

\section{Training Details}

All the hyperparameters are tuned on the validation set. For each of our main models (baseline, CIDEr and CIDEnt), we report the results on a $5$-avg-ensemble, where we run the model $5$ times with different initialization random seeds and take the average probabilities at each time step of the decoder during inference time. We use a fixed size step LSTM-RNN encoder-decoder, with encoder step size of $50$ and decoder step size of $16$. Each LSTM has a hidden size of $1024$. We use Inception-v4 features as video frame-level features. We use word embedding size of $512$. Also, we project down the $1536$-dim image features (Inception-v4) to $512$-dim. 

We apply dropout  to vertical connections as proposed in~\newcite{zaremba2014recurrent}, with a value $0.5$ and a gradient clip size of $10$. We use Adam optimizer~\cite{kingma2014adam} with a learning rate of $0.0001$ for baseline cross-entropy loss. All the trainable weights are initialized with a uniform distribution in the range $[-0.08,0.08]$. During the test time inference, we use beam search of size $5$. All our reward-based models use mixed loss optimization ~\cite{paulus2017deep,wu2016google}, where we train the model based on weighted ($\gamma$) combination of cross-entropy loss and reinforcement loss. For MSR-VTT dataset, we use $\gamma=0.9995$ for our CIDEr-RL model and $\gamma=0.9990$ for our CIDEnt-RL model. For YouTube2Text/MSVD dataset, we use $\gamma=0.9985$ for our CIDEr-RL model and $\gamma=0.9990$ and for our CIDEnt-RL model. The learning rate for the mixed-loss optimization is $1 \times 10^{-5}$ for MSR-VTT, and $1 \times 10^{-6}$ for YouTube2Text/MSVD.
The $\lambda$ hyperparameter in our CIDEnt reward formulation (see Sec. 4 in main paper) is roughly equal to the baseline cross-entropy model's score on that metric, i.e., $\lambda=0.45$ for MSR-VTT CIDEnt-RL model and $\lambda=0.75$ for YouTube2Text/MSVD CIDEnt-RL model.

\begin{figure}[t]
\centering
\includegraphics[width=0.98\linewidth]{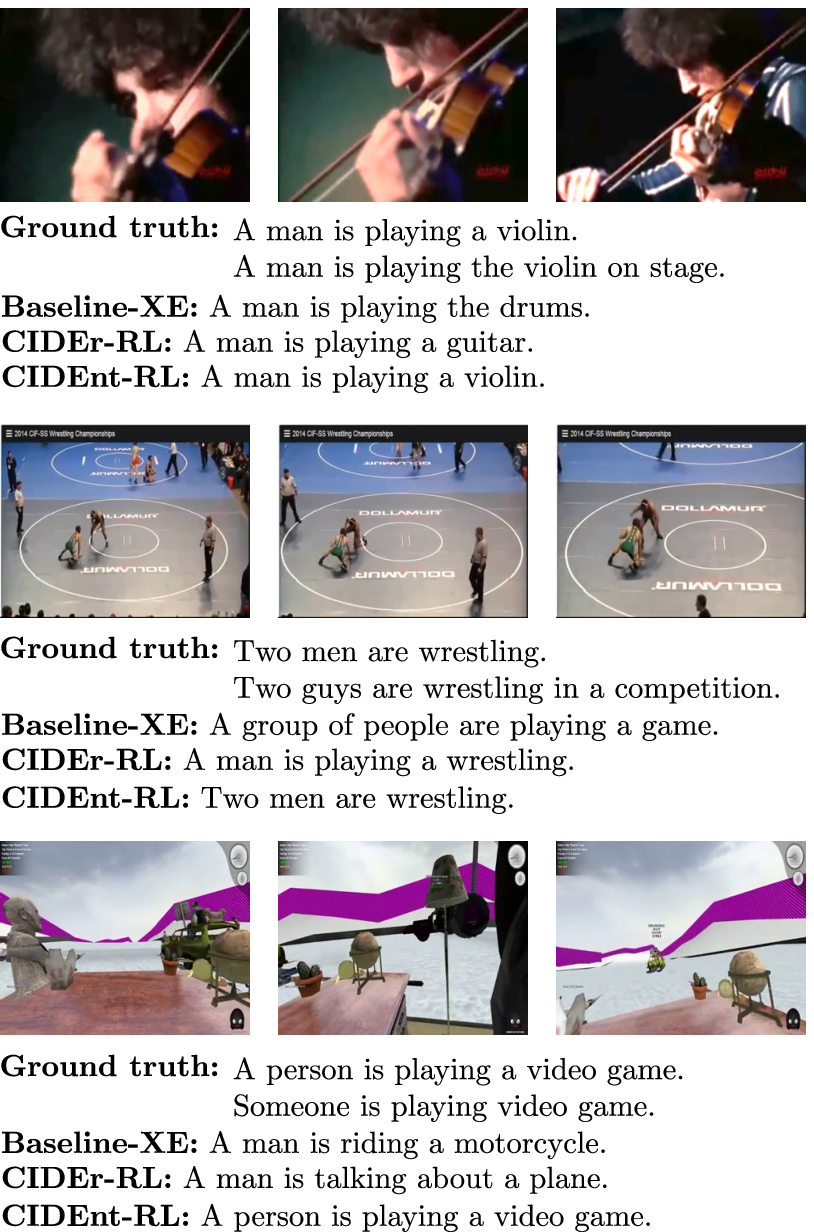}
\caption{Output examples where our CIDEnt-RL model produces better entailed captions than the phrase-matching CIDEr-RL model, which in turn is better than the baseline cross-entropy model.
}
\label{fig:analysis}
\end{figure}

\section{Analysis}
Figure~\ref{fig:analysis} shows several examples where our CIDEnt-reward model produces better entailed captions than the ones generated by the CIDEr-reward model. This is because the CIDEr-style captioning metrics achieve a high score even when the generation does not exactly entail the ground truth but is just a high phrase overlap. This can obviously cause issues by inserting a single wrong word such as a negation, contradiction, or wrong action/object. On the other hand, our entailment-enhanced CIDEnt score is only high when both CIDEr and the entailment classifier achieve high scores. The CIDEr-RL model, in turn, produces better captions than the baseline cross-entropy model, which is not aware of sentence-level matching at all.

%% file: main.bbl
\begin{thebibliography}{39}
\expandafter\ifx\csname natexlab\endcsname\relax\def\natexlab#1{#1}\fi

\bibitem[{Anderson et~al.(2016)Anderson, Fernando, Johnson, and
  Gould}]{anderson2016spice}
Peter Anderson, Basura Fernando, Mark Johnson, and Stephen Gould. 2016.
\newblock {SPICE}: Semantic propositional image caption evaluation.
\newblock In \emph{ECCV}, pages 382--398.

\bibitem[{Bahdanau et~al.(2015)Bahdanau, Cho, and Bengio}]{bahdanau2014neural}
Dzmitry Bahdanau, Kyunghyun Cho, and Yoshua Bengio. 2015.
\newblock Neural machine translation by jointly learning to align and
  translate.
\newblock In \emph{ICLR}.

\bibitem[{Bowman et~al.(2015)Bowman, Angeli, Potts, and
  Manning}]{bowman2015large}
Samuel~R Bowman, Gabor Angeli, Christopher Potts, and Christopher~D Manning.
  2015.
\newblock A large annotated corpus for learning natural language inference.
\newblock In \emph{EMNLP}.

\bibitem[{Chen and Dolan(2011)}]{chen2011collecting}
David~L Chen and William~B Dolan. 2011.
\newblock Collecting highly parallel data for paraphrase evaluation.
\newblock In \emph{Proceedings of the 49th Annual Meeting of the Association
  for Computational Linguistics: Human Language Technologies-Volume 1}, pages
  190--200. Association for Computational Linguistics.

\bibitem[{Chen et~al.(2015)Chen, Fang, Lin, Vedantam, Gupta, Doll{\'a}r, and
  Zitnick}]{chen2015microsoft}
Xinlei Chen, Hao Fang, Tsung-Yi Lin, Ramakrishna Vedantam, Saurabh Gupta, Piotr
  Doll{\'a}r, and C~Lawrence Zitnick. 2015.
\newblock Microsoft {COCO} captions: Data collection and evaluation server.
\newblock \emph{arXiv preprint arXiv:1504.00325}.

\bibitem[{Cheng et~al.(2016)Cheng, Dong, and Lapata}]{cheng2016long}
Jianpeng Cheng, Li~Dong, and Mirella Lapata. 2016.
\newblock Long short-term memory-networks for machine reading.
\newblock In \emph{EMNLP}.

\bibitem[{Dagan et~al.(2006)Dagan, Glickman, and Magnini}]{dagan2006pascal}
Ido Dagan, Oren Glickman, and Bernardo Magnini. 2006.
\newblock The {PASCAL} recognising textual entailment challenge.
\newblock In \emph{Machine learning challenges. evaluating predictive
  uncertainty, visual object classification, and recognising tectual
  entailment}, pages 177--190. Springer.

\bibitem[{Denkowski and Lavie(2014)}]{banerjee2005meteor}
Michael Denkowski and Alon Lavie. 2014.
\newblock Meteor universal: Language specific translation evaluation for any
  target language.
\newblock In \emph{EACL}.

\bibitem[{Efron and Tibshirani(1994)}]{efron1994introduction}
Bradley Efron and Robert~J Tibshirani. 1994.
\newblock \emph{An introduction to the bootstrap}.
\newblock CRC press.

\bibitem[{Elliott and Keller(2014)}]{elliott2014comparing}
Desmond Elliott and Frank Keller. 2014.
\newblock Comparing automatic evaluation measures for image description.
\newblock In \emph{ACL}, pages 452--457.

\bibitem[{Hodosh et~al.(2013)Hodosh, Young, and
  Hockenmaier}]{hodosh2013framing}
Micah Hodosh, Peter Young, and Julia Hockenmaier. 2013.
\newblock Framing image description as a ranking task: Data, models and
  evaluation metrics.
\newblock \emph{Journal of Artificial Intelligence Research}, 47:853--899.

\bibitem[{Jimenez et~al.(2014)Jimenez, Duenas, Baquero, Gelbukh, B{\'a}tiz, and
  Mendiz{\'a}bal}]{jimenez2014unal}
Sergio Jimenez, George Duenas, Julia Baquero, Alexander Gelbukh, Av~Juan~Dios
  B{\'a}tiz, and Av~Mendiz{\'a}bal. 2014.
\newblock {UNAL-NLP}: Combining soft cardinality features for semantic textual
  similarity, relatedness and entailment.
\newblock In \emph{In {SemEval}}, pages 732--742.

\bibitem[{Kingma and Ba(2015)}]{kingma2014adam}
Diederik Kingma and Jimmy Ba. 2015.
\newblock Adam: A method for stochastic optimization.
\newblock In \emph{ICLR}.

\bibitem[{Lai and Hockenmaier(2014)}]{lai2014illinois}
Alice Lai and Julia Hockenmaier. 2014.
\newblock Illinois-{LH}: A denotational and distributional approach to
  semantics.
\newblock \emph{Proc. SemEval}, 2:5.

\bibitem[{Lin(2004)}]{lin2004rouge}
Chin-Yew Lin. 2004.
\newblock {ROUGE}: A package for automatic evaluation of summaries.
\newblock In \emph{Text Summarization Branches Out: Proceedings of the ACL-04
  workshop}, volume~8.

\bibitem[{Liu et~al.(2016{\natexlab{a}})Liu, Lowe, Serban, Noseworthy, Charlin,
  and Pineau}]{liu2016not}
Chia-Wei Liu, Ryan Lowe, Iulian~V Serban, Michael Noseworthy, Laurent Charlin,
  and Joelle Pineau. 2016{\natexlab{a}}.
\newblock How not to evaluate your dialogue system: An empirical study of
  unsupervised evaluation metrics for dialogue response generation.
\newblock In \emph{EMNLP}.

\bibitem[{Liu et~al.(2016{\natexlab{b}})Liu, Zhu, Ye, Guadarrama, and
  Murphy}]{liu2016improved}
Siqi Liu, Zhenhai Zhu, Ning Ye, Sergio Guadarrama, and Kevin Murphy.
  2016{\natexlab{b}}.
\newblock Improved image captioning via policy gradient optimization of
  {SPIDE}r.
\newblock \emph{arXiv preprint arXiv:1612.00370}.

\bibitem[{Noreen(1989)}]{noreen1989computer}
Eric~W Noreen. 1989.
\newblock \emph{Computer-intensive methods for testing hypotheses}.
\newblock Wiley New York.

\bibitem[{Pan et~al.(2016{\natexlab{a}})Pan, Xu, Yang, Wu, and
  Zhuang}]{pan2015hierarchical}
Pingbo Pan, Zhongwen Xu, Yi~Yang, Fei Wu, and Yueting Zhuang.
  2016{\natexlab{a}}.
\newblock Hierarchical recurrent neural encoder for video representation with
  application to captioning.
\newblock In \emph{Proceedings of the IEEE Conference on Computer Vision and
  Pattern Recognition}, pages 1029--1038.

\bibitem[{Pan et~al.(2016{\natexlab{b}})Pan, Mei, Yao, Li, and
  Rui}]{pan2015jointly}
Yingwei Pan, Tao Mei, Ting Yao, Houqiang Li, and Yong Rui. 2016{\natexlab{b}}.
\newblock Jointly modeling embedding and translation to bridge video and
  language.
\newblock In \emph{Proceedings of the IEEE Conference on Computer Vision and
  Pattern Recognition}, pages 4594--4602.

\bibitem[{Papineni et~al.(2002)Papineni, Roukos, Ward, and
  Zhu}]{papineni2002bleu}
Kishore Papineni, Salim Roukos, Todd Ward, and Wei-Jing Zhu. 2002.
\newblock {BLEU}: a method for automatic evaluation of machine translation.
\newblock In \emph{ACL}, pages 311--318.

\bibitem[{Parikh et~al.(2016)Parikh, T{\"a}ckstr{\"o}m, Das, and
  Uszkoreit}]{parikh2016decomposable}
Ankur~P Parikh, Oscar T{\"a}ckstr{\"o}m, Dipanjan Das, and Jakob Uszkoreit.
  2016.
\newblock A decomposable attention model for natural language inference.
\newblock In \emph{EMNLP}.

\bibitem[{Pasunuru and Bansal(2017)}]{pasunuru2017multitask}
Ramakanth Pasunuru and Mohit Bansal. 2017.
\newblock Multi-task video captioning with video and entailment generation.
\newblock In \emph{Proceedings of ACL}.

\bibitem[{Paulus et~al.(2017)Paulus, Xiong, and Socher}]{paulus2017deep}
Romain Paulus, Caiming Xiong, and Richard Socher. 2017.
\newblock A deep reinforced model for abstractive summarization.
\newblock \emph{arXiv preprint arXiv:1705.04304}.

\bibitem[{Ranzato et~al.(2016)Ranzato, Chopra, Auli, and
  Zaremba}]{ranzato2015sequence}
Marc'Aurelio Ranzato, Sumit Chopra, Michael Auli, and Wojciech Zaremba. 2016.
\newblock Sequence level training with recurrent neural networks.
\newblock In \emph{ICLR}.

\bibitem[{Rennie et~al.(2016)Rennie, Marcheret, Mroueh, Ross, and
  Goel}]{rennie2016self}
Steven~J Rennie, Etienne Marcheret, Youssef Mroueh, Jarret Ross, and Vaibhava
  Goel. 2016.
\newblock Self-critical sequence training for image captioning.
\newblock \emph{arXiv preprint arXiv:1612.00563}.

\bibitem[{Rockt{\"a}schel et~al.(2016)Rockt{\"a}schel, Grefenstette, Hermann,
  Ko{\v{c}}isk{\`y}, and Blunsom}]{rocktaschel2015reasoning}
Tim Rockt{\"a}schel, Edward Grefenstette, Karl~Moritz Hermann, Tom{\'a}{\v{s}}
  Ko{\v{c}}isk{\`y}, and Phil Blunsom. 2016.
\newblock Reasoning about entailment with neural attention.
\newblock In \emph{ICLR}.

\bibitem[{Szegedy et~al.(2016)Szegedy, Ioffe, and
  Vanhoucke}]{szegedy2016inception}
Christian Szegedy, Sergey Ioffe, and Vincent Vanhoucke. 2016.
\newblock Inception-v4, inception-resnet and the impact of residual connections
  on learning.
\newblock In \emph{CoRR}.

\bibitem[{Vedantam et~al.(2015)Vedantam, Lawrence~Zitnick, and
  Parikh}]{vedantam2015cider}
Ramakrishna Vedantam, C~Lawrence~Zitnick, and Devi Parikh. 2015.
\newblock {CIDE}r: Consensus-based image description evaluation.
\newblock In \emph{CVPR}, pages 4566--4575.

\bibitem[{Venugopalan et~al.(2016)Venugopalan, Hendricks, Mooney, and
  Saenko}]{venugopalan2016improving}
Subhashini Venugopalan, Lisa~Anne Hendricks, Raymond Mooney, and Kate Saenko.
  2016.
\newblock Improving lstm-based video description with linguistic knowledge
  mined from text.
\newblock In \emph{EMNLP}.

\bibitem[{Venugopalan et~al.(2015{\natexlab{a}})Venugopalan, Rohrbach, Donahue,
  Mooney, Darrell, and Saenko}]{venugopalan2015sequence}
Subhashini Venugopalan, Marcus Rohrbach, Jeffrey Donahue, Raymond Mooney,
  Trevor Darrell, and Kate Saenko. 2015{\natexlab{a}}.
\newblock Sequence to sequence-video to text.
\newblock In \emph{CVPR}, pages 4534--4542.

\bibitem[{Venugopalan et~al.(2015{\natexlab{b}})Venugopalan, Xu, Donahue,
  Rohrbach, Mooney, and Saenko}]{venugopalan2014translating}
Subhashini Venugopalan, Huijuan Xu, Jeff Donahue, Marcus Rohrbach, Raymond
  Mooney, and Kate Saenko. 2015{\natexlab{b}}.
\newblock Translating videos to natural language using deep recurrent neural
  networks.
\newblock In \emph{{NAACL} {HLT}}.

\bibitem[{Williams et~al.(2017)Williams, Nangia, and
  Bowman}]{williams2017broad}
Adina Williams, Nikita Nangia, and Samuel~R Bowman. 2017.
\newblock A broad-coverage challenge corpus for sentence understanding through
  inference.
\newblock \emph{arXiv preprint arXiv:1704.05426}.

\bibitem[{Williams(1992)}]{williams1992simple}
Ronald~J Williams. 1992.
\newblock Simple statistical gradient-following algorithms for connectionist
  reinforcement learning.
\newblock \emph{Machine learning}, 8(3-4):229--256.

\bibitem[{Wu et~al.(2016)Wu, Schuster, Chen, Le, Norouzi, Macherey, Krikun,
  Cao, Gao, Macherey et~al.}]{wu2016google}
Yonghui Wu, Mike Schuster, Zhifeng Chen, Quoc~V Le, Mohammad Norouzi, Wolfgang
  Macherey, Maxim Krikun, Yuan Cao, Qin Gao, Klaus Macherey, et~al. 2016.
\newblock Google's neural machine translation system: Bridging the gap between
  human and machine translation.
\newblock \emph{arXiv preprint arXiv:1609.08144}.

\bibitem[{Xu et~al.(2016)Xu, Mei, Yao, and Rui}]{xu2016msr}
Jun Xu, Tao Mei, Ting Yao, and Yong Rui. 2016.
\newblock {MSR-VTT}: A large video description dataset for bridging video and
  language.
\newblock In \emph{CVPR}, pages 5288--5296.

\bibitem[{Yao et~al.(2015)Yao, Torabi, Cho, Ballas, Pal, Larochelle, and
  Courville}]{yao2015describing}
Li~Yao, Atousa Torabi, Kyunghyun Cho, Nicolas Ballas, Christopher Pal, Hugo
  Larochelle, and Aaron Courville. 2015.
\newblock Describing videos by exploiting temporal structure.
\newblock In \emph{CVPR}, pages 4507--4515.

\bibitem[{Zaremba and Sutskever(2015)}]{zaremba2015reinforcement}
Wojciech Zaremba and Ilya Sutskever. 2015.
\newblock Reinforcement learning neural turing machines.
\newblock \emph{arXiv preprint arXiv:1505.00521}, 362.

\bibitem[{Zaremba et~al.(2014)Zaremba, Sutskever, and
  Vinyals}]{zaremba2014recurrent}
Wojciech Zaremba, Ilya Sutskever, and Oriol Vinyals. 2014.
\newblock Recurrent neural network regularization.
\newblock \emph{arXiv preprint arXiv:1409.2329}.

\end{thebibliography}
